\documentclass[runningheads]{llncs}

\usepackage{accv}

\usepackage{accvabbrv}
\usepackage{float}
\usepackage[normalem]{ulem}
\usepackage{graphicx}
\usepackage{booktabs}
\usepackage[table]{xcolor}
\usepackage{multirow}
\usepackage{graphicx} 
\usepackage{ulem}
\usepackage{amsmath}
\usepackage[accsupp]{axessibility}  
\usepackage{multirow}
\usepackage[table,xcdraw]{xcolor}
\usepackage[normalem]{ulem}
\useunder{\uline}{\ul}{}

\usepackage[pagebackref,breaklinks,colorlinks,citecolor=accvblue]{hyperref}

\usepackage{orcidlink}

\begin{document}

\title{M2P-AD: Memory-to-Prototype Learning \\ with Boundary-aware Score Refinement \\for 3D Anomaly Detection} 

\titlerunning{Memory-to-Prototype Anomaly Detection}


\author{Seyoung Jeong\inst{1} \and
Jong Pil Yun\inst{2} \and
Sang Jun Lee\inst{1,*}}

\authorrunning{S.~Jeong et al.}

\institute{Jeonbuk National University, Jeonju, Republic of Korea\\
\email{\{wjdtpdud, sj.lee\}@jbnu.ac.kr}\\
\and
Manufacturing AI Research Center, Korea Institute of Industrial Technology (KITECH), Incheon, Republic of Korea\\
\email{rebirth@kitech.re.kr}}




\maketitle

\begin{abstract}
  3D anomaly detection has recently emerged as an important research topic in computer vision.
  Although existing methods have achieved high performance, excessive anomaly responses in normal regions and false positives near object boundaries remain unresolved challenges.  
  To address these challenges, we propose a novel 3D anomaly detection model, Memory-to-Prototype Anomaly Detection (M2P-AD), which effectively models the distribution of normal features while suppressing excessive anomaly scores in normal regions and false positives near object boundaries.
  Specifically, we introduce a Memory-to-Prototype (M2P) module that learns representative prototypes from normal feature embeddings to preserve important structural information of objects.
  In addition, a Boundary extraction (BE) module is integrated to identify object boundaries, and a Boundary-aware score refinement (BSR) strategy is applied to recalibrate anomaly scores by incorporating boundary characteristics.
  The proposed method is evaluated on Real3D-AD, Anomaly-ShapeNet, and MulSen-AD, achieving state-of-the-art performance.
  Qualitative results demonstrate that excessive anomaly scores in normal regions are reduced and false positives near object boundaries are suppressed, resulting in more accurate and stable anomaly localization.
  The results indicate that the proposed approach enables more reliable 3D anomaly detection and provides a robust solution applicable to real-world industrial environments.
  \keywords{3D anomaly detection \and industrial applications \and memory bank \and prototype learning}
\end{abstract}

\section{Introduction}
\label{sec:intro}
Anomaly detection is a fundamental technique for ensuring operational safety and maintaining product quality in manufacturing industries.
Particularly in real-world industrial environments, where defect occurrences are rare and precise labeling requires time and cost, unsupervised anomaly detection methods that utilize only normal data have attracted increasing attention.
Existing unsupervised anomaly detection\cite{bae2023pni, lei2023pyramidflow} research has primarily focused on 2D approaches based on RGB images, achieving meaningful performance improvements across various defect types.
However, 2D-based methods rely solely on appearance information and therefore remain limited in accurately capturing structural defects, such as geometric variations or subtle shape deformations.
To effectively capture geometric and structural characteristics, recent research has increasingly focused on unsupervised anomaly detection based on 3D point clouds\cite{zhou2024r3d}.

The inherent irregularity and sparsity of point cloud data pose significant challenges for modeling structural relationships and estimating the normal distribution. 
Accordingly, prior 3D anomaly detection methods\cite{liu2023real3d, chen2023easynet} have predominantly adopted memory bank–based frameworks to model normal representations using handcrafted descriptors or pretrained neural networks and to detect anomalies through distance-based similarity evaluation against test samples.
For example, ISMP\cite{liang2025look} further improves performance by capturing intrinsic structural information from point clouds and modeling the complementary relationship between internal and external feature representations.

\begin{figure}[t]
  \centering
  \includegraphics[width=0.9\linewidth]{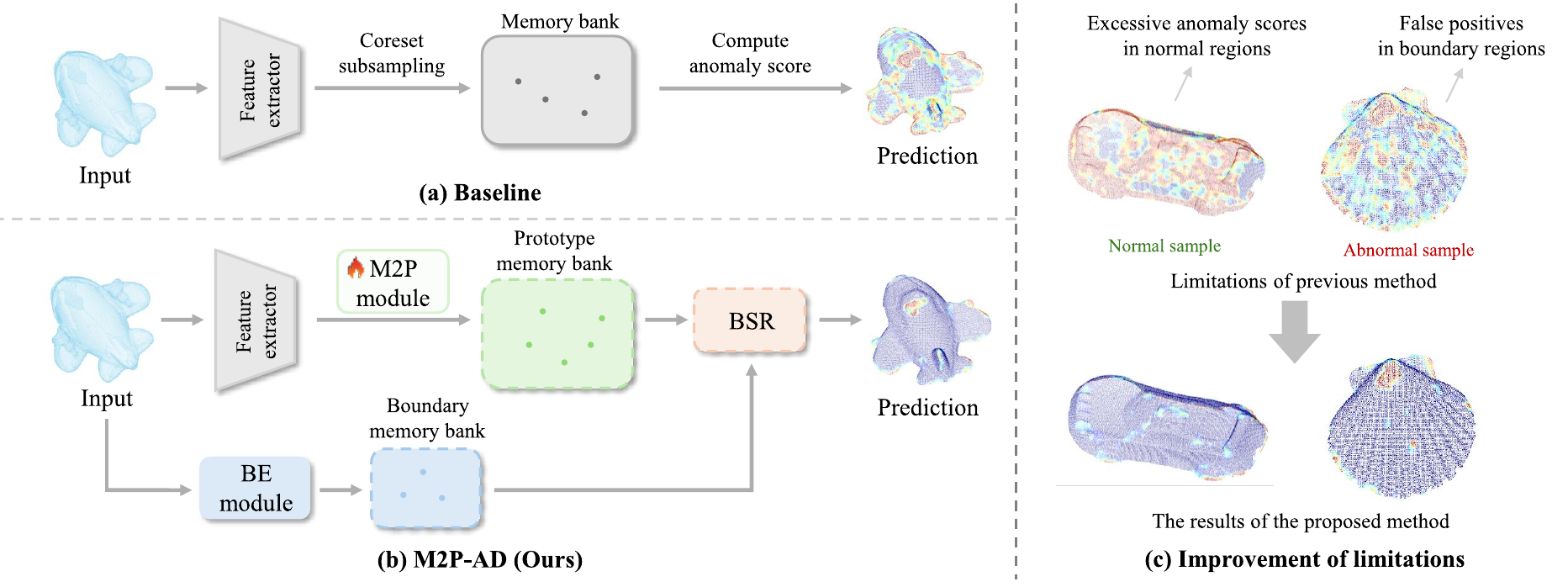}
  \caption{Comparison between the baseline and the proposed framework. (a) Baseline architecture. (b) Proposed architecture incorporating the M2P module and the BSR. (c) Qualitative comparison showing the improvements achieved by the proposed method.
  }
  \label{fig:seyoung1}
\end{figure}

Despite recent advances in 3D anomaly detection, fundamental challenges remain in achieving precise and reliable anomaly detection.
As presented in \cref{fig:seyoung1} (a), most memory bank–based methods adopt distance-based coreset subsampling to enhance memory efficiency. 
However, this process fails to preserve structurally important patches from informative regions of normal objects, leading to a loss of representativeness in the memory bank.
As a result, as presented in \cref{fig:seyoung1} (c), even normal regions fail to find appropriate correspondences in the memory bank during inference, resulting in excessively high anomaly scores.
Furthermore, structural boundary regions often receive disproportionately high anomaly scores, despite being part of the normal object structure.
The existing methods apply uniform distance-based criteria without accounting for the inherent characteristics of boundary regions, leading to frequent false positives (FP) and significantly reducing the reliability of anomaly detection results.

To address excessive anomaly scores in normal regions and frequent FP in structural boundary regions, this paper proposes Memory-to-Prototype Anomaly Detection (M2P-AD), a novel memory bank–based 3D anomaly detection framework.
An overview of the proposed architecture is presented in \cref{fig:seyoung1} (b).
We propose Memory-to-Prototype (M2P), a module that learns representative prototype embeddings from normal feature embeddings to model the normal data distribution.
Specifically, normal feature embeddings are partitioned into 128 semantic clusters, and for each cluster, a representative prototype embedding is learned by fusing Euclidean distance and cosine similarity.
The M2P module effectively mitigates the loss of structurally important normal patches introduced by distance-based coreset subsampling.
In addition, to alleviate FP concentrated in structural boundary regions, the framework incorporates a Boundary extraction (BE) module to identify boundary regions and introduces a Boundary-aware score refinement (BSR) strategy to recalibrate anomaly scores based on boundary characteristics.
Consequently, the proposed M2P-AD framework preserves critical structural representations of the normal distribution while effectively suppressing boundary-induced anomaly responses, resulting in more reliable 3D anomaly detection.

Our contributions are summarized as follows:
\begin{itemize}
\item We propose M2P-AD, a novel framework for reliable 3D anomaly detection.
\item We propose an M2P module that learns representative prototypes to model the normal distribution and preserve structurally important normal-region information.
\item We propose a BE module for boundary identification and a BSR strategy for suppressing boundary-induced FP.
\item Experiments demonstrate that the proposed method achieves state-of-the-art performance on Real3D-AD, Anomaly-ShapeNet, and MulSen-AD.
\end{itemize}

\section{Related work}

\subsection{2D Anomaly detection}
Unsupervised anomaly detection on RGB images has made substantial progress in recent years~\cite{wyatt2022anoddpm, he2024diffusion}.
Existing 2D methods can be broadly categorized into feature embedding–based\cite{deng2022anomaly, rudolph2023asymmetric} and reconstruction-based\cite{fuvcka2024transfusion, yao2024glad} approaches.
Feature embedding–based methods leverage pretrained models to extract feature embeddings from normal samples and detect anomalies by measuring the deviation of test features from the normal data distribution.
As a representative example, PatchCore\cite{roth2022towards} stores feature embeddings of normal samples in a memory bank to model the normal distribution and computes anomaly scores based on distance-based similarity between test samples and the memory bank.
AST\cite{rudolph2023asymmetric} adopts an asymmetric student–teacher architecture to learn normal representations and alleviate performance degradation caused by domain shifts.
Reconstruction-based approaches utilize autoencoders~\cite{bergmann2018improving,hong2020latent} or generative adversarial networks (GANs)~\cite{carrara2021combining} to learn normal representations and compute anomaly scores based on the discrepancy between the input and the reconstruction result.
DRAEM\cite{zavrtanik2021draem} injects synthetically generated anomalies into normal images and trains the model to restore them, thereby improving discrimination between normal and abnormal patterns.
However, 2D-based approaches rely solely on appearance and color information, making them limited in capturing structural defects and geometric variations, which has led to increasing interest in 3D anomaly detection methods that leverage depth and geometric information.

\subsection{3D Anomaly detection}
The introduction of dedicated benchmarks for point cloud anomaly detection, such as Real3D-AD~\cite{liu2023real3d} and Anomaly-ShapeNet~\cite{li2024towards}, has accelerated research on 3D point cloud–based anomaly detection methods.
Reg3D-AD\cite{liu2023real3d} leverages a pretrained PointMAE to extract features from normal samples and proposes a memory bank–based framework that preserves both local and global representations.
Reg2Inv\cite{yu2025registration} introduces a Rotation-Invariant Feature Extractor and a Registration-Induced Anomaly Detection strategy to address spatial alignment errors in point clouds.
MVP-PCLIP\cite{cheng2025toward} projects point cloud data into multi-view depth images and leverages pretrained vision–language models to improve anomaly detection and generalization across product categories.
While these methods have improved 3D anomaly detection performance, existing approaches still struggle to fully leverage normal samples.
Therefore, we propose M2P-AD, a robust 3D anomaly detection framework that more effectively models the normal distribution and improves detection reliability.

\section{Method}
\subsection{Framework overview}
\begin{figure}[t]
  \centering
  \includegraphics[width=1.0\linewidth]{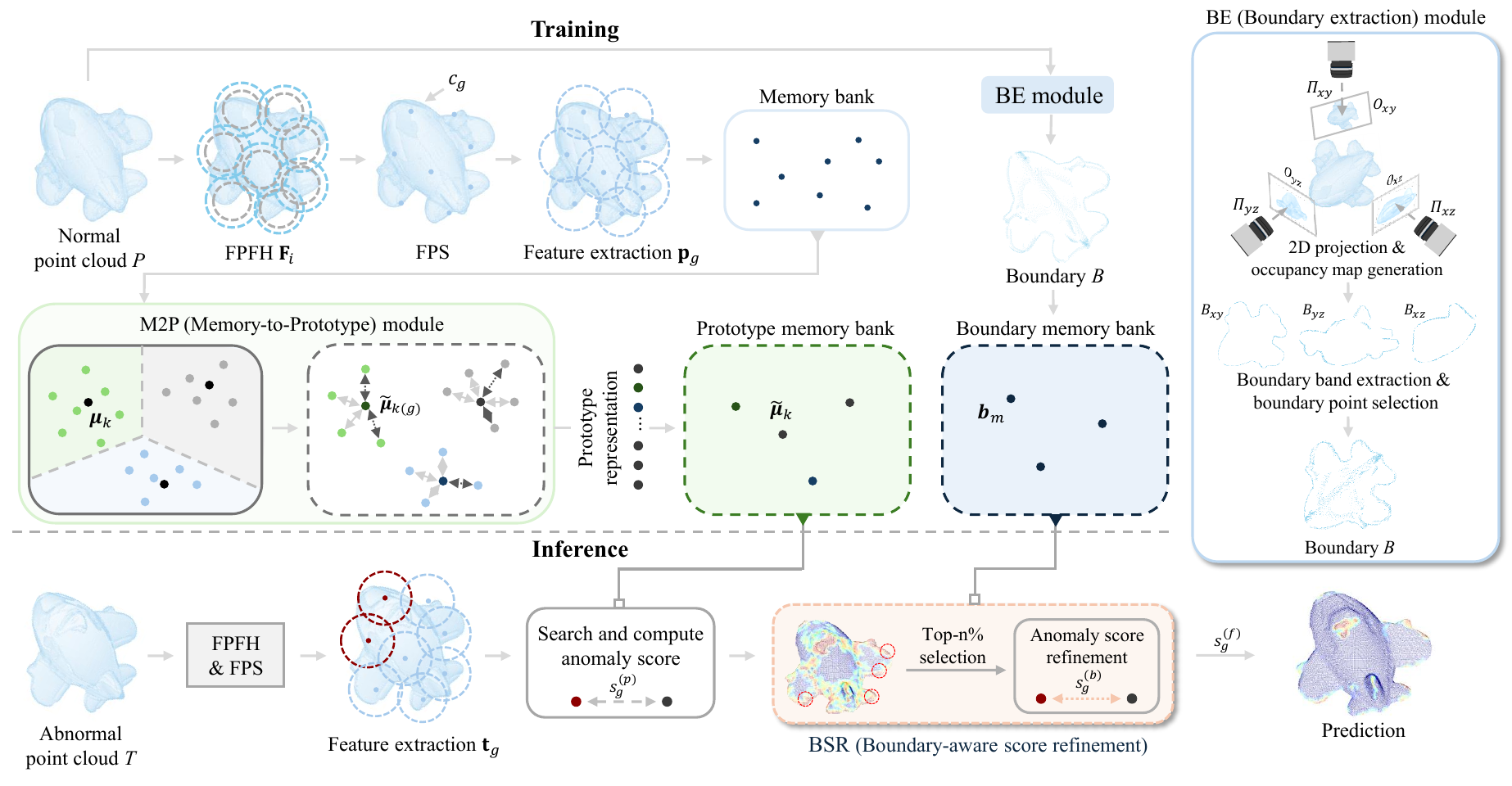}
  \caption{Overall architecture of the Memory-to-Prototype Anomaly Detection.}
  \label{fig:seyoung2}
\end{figure}

In this paper, we propose M2P-AD, a 3D anomaly detection framework to address excessive anomaly scores in normal regions and frequent FP in structural boundary regions.
The overall architecture of the proposed method is presented in \cref{fig:seyoung2}.
We extract patch-level embeddings from normal data and use the M2P module to build a memory bank consisting of representative prototype vectors, thereby preserving representative structures of the normal distribution.
Simultaneously, the point cloud is projected into the image space to extract structural boundaries, and a boundary memory bank is constructed by matching the corresponding boundary patches.
During inference, anomaly scores are initially computed based on the prototype memory bank. The top-$n\%$ highest-scoring patches are subsequently selected and further refined through comparison with the boundary memory bank.
The proposed framework consists of two main components: the M2P module that learns representative prototypes to model the normal distribution, and a boundary-aware refinement component composed of the BE module and the BSR strategy.
Detailed descriptions of each component are presented in the following sections.

\subsection{Feature extraction}
We extract patch-wise geometric features to encode the local structural characteristics of the input point cloud.
Given a normal point cloud $P = \{p_i\}_{i=1}^{I}$, we compute a Fast Point Feature Histogram (FPFH) descriptor for each point $p_i$ to capture local geometric structure.
We first estimate surface normals $\mathbf{n}_i$ and then extract the corresponding FPFH features $\mathbf{f}_i$ within the neighborhood $\mathcal{N}(p_i)$.
\begin{align}
\mathbf{f}_i = \mathrm{FPFH}\big(p_i, \mathcal{N}(p_i), \mathbf{n}_i\big).
\end{align}
To capture broader geometric context, FPFH are recomputed using multi-scale neighborhood radius $r \in \{40, 80, 120\}$.
The resulting features are concatenated along the channel dimension to form a multi-scale feature $\mathbf{F}_i$ for each point.

Patch-level representations are then constructed by selecting $G$ representative centers $\{c_g\}_{g=1}^{G}$ using Farthest Point Sampling (FPS), which provides spatially well-distributed coverage over the entire point cloud, where $G=2048$.
Specifically, for each center $c_g$, a local neighborhood is defined using k-nearest neighbors, and the corresponding point descriptors are aggregated via mean pooling to produce a patch embedding.

\begin{align}
\mathbf{p}_g
=
\frac{1}{|\mathcal{N}(c_g)|}
\sum_{p_i \in \mathcal{N}(c_g)}
\mathbf{F}_i ,
\end{align}
where $|\mathcal{N}(c_g)|$ represents the number of neighboring points and is set to 128, and $\mathbf{p}_g$ denotes the resulting patch embedding that captures the geometric characteristics of the local neighborhood centered at $c_g$.
The extracted patch embeddings are then used to learn representative prototype vectors in the M2P module.

\subsection{Memory-to-Prototype module}
\label{sec:m2p}
We propose the M2P module to compactly model the normal feature distribution while preserving important information from normal regions.
To group similar patch embeddings and identify representative structures of the normal distribution, we apply K-means clustering to all extracted patch embeddings $\mathbf{p}_g$, partitioning them into $K$ clusters $\{\mathcal{C}_k\}_{k=1}^{K}$, where $K$ is set to 128.
To construct a compact memory bank while preserving structurally important normal-region information, a representative prototype vector is defined for each cluster.
The initial prototype vector representing each cluster is defined as the centroid $\boldsymbol{\mu}_k$.
Each patch embedding $\mathbf{p}_g$ is assigned to its nearest centroid based on Euclidean distance, forming a cluster $\mathcal{C}_k$  defined as:

\begin{align}
\mathcal{C}_k
=
\left\{
\mathbf{p}_g \ \bigg|\ 
k = \underset{j \in \{1,\ldots,K\}}{\operatorname{argmin}}
\left\| \mathbf{p}_g - \boldsymbol{\mu}_j \right\|_2
\right\}.
\end{align}

The prototype vectors are further refined via optimization, resulting in refined prototypes $\tilde{\boldsymbol{\mu}}_k$.
To alleviate the limited representativeness caused by relying on a single distance criterion, we additionally consider the angular consistency between each prototype and its assigned patch embeddings.
Specifically, we minimize the Euclidean distance between each patch embedding and its assigned prototype, with an additional cosine similarity–based angular loss:
\begin{align}
\mathcal{L}
=
\frac{1}{G}
\sum_{g=1}^{G}
\left(
\left\|
\mathbf{p}_g - \tilde{\boldsymbol{\mu}}_{k(g)}
\right\|_2^2
+
\left(
1 -
\frac{
\mathbf{p}_g^\top \tilde{\boldsymbol{\mu}}_{k(g)}
}{
\|\mathbf{p}_g\|_2 \, \|\tilde{\boldsymbol{\mu}}_{k(g)}\|_2
}
\right)
\right),
\end{align}
where $\tilde{\boldsymbol{\mu}}_{k(g)}$ is the refined prototype assigned to $\mathbf{p}_g$, and $k(g)$ indicates the index of the nearest prototype.
This optimization enables each prototype to effectively model the underlying normal feature distribution of its assigned cluster.
As a result, only the learned prototype vectors are retained to form the final prototype memory bank.

\subsection{Boundary extraction module}
\label{sec:be}
We propose to extract boundary patch features to construct a boundary memory bank, which suppresses frequent FP near object boundaries.
Specifically, the point cloud $P$ is projected onto the three canonical planes $xy$, $xz$, and $yz$ to generate 2D occupancy maps that preserve the geometric extent of the object from different viewpoints. 
The projection operator onto plane $\alpha \in \{xy,xz,yz\}$ is denoted by $\Pi_\alpha$.
The projection of each point $p_i = (x_i, y_i, z_i)$ onto the three planes is defined as:
\begin{align}
\Pi_{xy}(p_i) = (x_i, y_i), \Pi_{xz}(p_i) = (x_i, z_i), \Pi_{yz}(p_i) = (y_i, z_i).
\end{align}
Based on these projections, a binary occupancy map $O_\alpha(u,v)$ is generated for each plane $\alpha$, where $(u,v)$ denotes a 2D pixel location:
\begin{align}
O_{\alpha}(u,v)
=
\begin{cases}
1, & \text{if } \exists\, p_i \in P \text{ such that } \Pi_{\alpha}(p_i) = (u,v),\\
0, & \text{otherwise}.
\end{cases}
\end{align}
To obtain a robust representation of object boundaries and avoid relying on a single-pixel boundary, we compute a distance transform $D_\alpha(u,v)$ for each occupancy map $O_\alpha$ to measure the Euclidean distance from each interior pixel to the boundary $\partial O_\alpha$ of the occupied region:
\begin{align}
D_{\alpha}(u,v)
&=
\min_{(x,y)\in \partial O_{\alpha}}
\| (u,v) - (x,y) \|_2.
\end{align}
The boundary band $B_\alpha$ is then defined as the set of occupied pixels whose distance to the boundary does not exceed a predefined threshold $\tau$:
\begin{align}
B_{\alpha}
&=
\left\{
(u,v)
\;\middle|\;
O_{\alpha}(u,v)=1,\;
D_{\alpha}(u,v) \le \tau
\right\},
\end{align}
where $\tau$ controls the width of the boundary band and is set to 3 pixels.

For each boundary pixel $(u,v) \in B_\alpha$, the corresponding 3D surface point is selected as
\begin{align}
p^{*}_{\alpha}(u,v)
=
\underset{\{p_i| \Pi_{\alpha}(p_i)=(u,v)\}}{\operatorname{argmax}}
d_{\alpha}(p_i).
\end{align}
where $d_\alpha(p_i)$ denotes the depth coordinate of point $p_i$ along the projection axis associated with plane $\alpha$, such that the outermost surface point along the projection direction is selected when multiple points project onto the same pixel.
The union of boundary points obtained from all three projections forms the final set of 3D boundary points for the point cloud $P$ is denoted as $B$:

\begin{align}
B
=
\bigcup_{\alpha \in \{xy,xz,yz\}}
\left\{
p^{*}_{\alpha}(u,v)
\;\middle|\;
(u,v) \in B_{\alpha}
\right\}.
\end{align} 
For each boundary point $p_i \in B$, the patch embedding corresponding to the nearest patch center $c_g$ is selected and represented as $\boldsymbol{b}_m$. 
Boundary patch embeddings are aggregated across all normal training samples to construct the boundary memory bank, which models object boundary structures and mitigates FP near boundary regions during inference.

\subsection{Boundary-aware score refinement}
\label{sec:bsr}
To suppress boundary-induced FP, we propose a BSR strategy that refines anomaly scores based on the boundary memory bank.
Specifically, the initial anomaly score $s_g^{(p)}$ for the $g$-th test patch feature $\mathbf{t}_g$ is computed as the minimum distance to the prototype vectors $\{\tilde{\boldsymbol{\mu}}_k\}_{k=1}^{K}$ representing normal regions:
\begin{align}
s_g^{(p)}
=
\min_{k=1,\ldots,K}
\left\|
\mathbf{t}_g - \tilde{\boldsymbol{\mu}}_{k}
\right\|_2 .
\end{align}
To distinguish FP near object boundaries from genuine anomalies, only the top-$n\%$ patches with the highest anomaly scores are selected as candidates for boundary-aware refinement.
In our implementation, n is set to 20.
For these candidate patches, the boundary-aware score $s_g^{(b)}$ is computed as the minimum distance between the test patch feature $\mathbf{t}_g$ and the boundary patch embeddings $\{\boldsymbol{b}_m\}_{m=1}^{M}$ stored in the boundary memory bank:
\begin{align}
s_g^{(b)}
=
\min_{m=1,\ldots,M}
\left\|
\mathbf{t}_g - \boldsymbol{b}_m
\right\|_2 .
\end{align}
The final anomaly score $s_g^{(f)}$ is computed as the minimum of $s_g^{(p)}$ and $s_g^{(b)}$, suppressing FP near boundaries while retaining high scores for true anomalies.
\begin{align}
s_g^{(f)} = \min \left( s_g^{(p)},\, s_g^{(b)} \right).
\end{align}
The proposed boundary-aware score refinement effectively reduces FP in geometrically complex regions while preserving sensitivity to true anomalies.

\section{Experiments}

\subsection{Experiment settings}
\noindent\textbf{Datasets.}
We evaluate the proposed method on Real3D-AD\cite{liu2023real3d}, Anomaly-ShapeNet\cite{li2024towards}, and MulSen-AD\cite{li2025multi}, which are benchmarks for 3D anomaly detection.
Real3D-AD is a high-resolution real-world dataset with 12 object categories, each containing 4 normal training samples and 100 test samples.
Anomaly-ShapeNet is a synthetic dataset comprising 40 categories and 1,600 anomalous samples.
The training set for each class contains four normal samples, while the test set consists of both normal and anomalous samples.
MulSen-AD is a real-world multimodal dataset, from which we use only point clouds, containing 15 categories, 1,391 training samples, and 644 test samples.

\noindent\textbf{Evaluation metrics and experimental environments.}
We evaluate performance using the Area Under the Receiver Operating Characteristic Curve (AUROC). 
Pixel-level AUROC (P-AUROC) evaluates point-wise localization accuracy, while object-level AUROC (O-AUROC) measures object-level detection accuracy.
All experiments were conducted on an NVIDIA H100 GPU and trained for 1000 epochs with a learning rate of 0.001 using the Adam optimizer.

\begin{table}[t]
\caption{Comparison of quantitative results on the Real3D-AD. The best performance is in \textbf {bold}, and the second best is {\ul underlined}.}
\label{tab:real3d-ad}
\centering
\setlength{\tabcolsep}{4.5pt}
\resizebox{\linewidth}{!}{%
\begin{tabular}{c|l|cccccccccccc|c}
\hline
& Method & Airplane & Candybar & Car & Chicken & Diamond & Duck & Fish & Gemstone & Seahorse & Shell & Starfish & Toffees & Mean \\ \hline
 & IMRNet\cite{li2024towards}  & 76.2 & 75.5 & 71.1 & 78.0 & 90.5 & 51.7 & 88.0 & 67.4 & 60.4 & 66.5 & 67.4 & 77.4 & 72.5 \\ 
 & ISMP\cite{liang2025look}  & \textbf{85.8} & {\ul 85.2} & 73.1 & 71.4 & 94.8 & 71.2 & {\ul 94.5} & 46.8 & 72.9 & 62.3 & 66.0 & 84.2 & 76.7 \\ 
 & PO3AD\cite{ye2025po3ad}  & 80.4 & 78.5 & 65.4 & 68.6 & 80.1 & 82.0 & 85.9 & {\ul 69.3} & 75.6 & {\ul 80.0} & 75.8 & 77.1 & 76.5 \\ 
 & MC3D-AD\cite{cheng2025mc3d}  & {\ul 85.0} & 83.0 & {\ul 74.9} & 71.5 & 95.5 & \textbf{83.1} & 86.5 & 56.0 & 71.6 & \textbf{80.3} & {\ul 76.6} & 73.8 & 78.2 \\ 
 & Simple3D\cite{cheng2026towards}  & 76.5 & 65.1 & \textbf{98.1} & {\ul 82.6} & \textbf{100.0} & 77.8 & 91.2 & \textbf{70.4} &\textbf{93.0} & 51.4 & 69.6 & {\ul 88.8} & {\ul 80.4} \\ 
\multirow{-6}{*}{\rotatebox{90}{O-AUROC}} & \cellcolor[HTML]{D9D9D9}Ours
& \cellcolor[HTML]{D9D9D9}69.8
& \cellcolor[HTML]{D9D9D9}\textbf{93.2}
& \cellcolor[HTML]{D9D9D9}63.9
& \cellcolor[HTML]{D9D9D9}\textbf{86.4}
& \cellcolor[HTML]{D9D9D9}{\ul 99.1}
& \cellcolor[HTML]{D9D9D9}{\ul 82.4}
& \cellcolor[HTML]{D9D9D9}\textbf{97.2}
& \cellcolor[HTML]{D9D9D9}68.5
& \cellcolor[HTML]{D9D9D9}{\ul 91.1}
& \cellcolor[HTML]{D9D9D9}63.1
& \cellcolor[HTML]{D9D9D9}\textbf{82.9}
& \cellcolor[HTML]{D9D9D9}\textbf{92.2}
& \cellcolor[HTML]{D9D9D9}\textbf{82.4} \\ \hline

 & IMRNet\cite{li2024towards}  & - & - & - & - & - & - & - & - & - & - & - & - & - \\ 
 & ISMP\cite{liang2025look}  & 75.3 & 90.7 & 83.6 & 79.8 & 92.6 & 87.6 & 88.6 & 85.7 & 81.3 & {\ul 83.9} & 64.1 & 89.5 & 83.6 \\ 
 & PO3AD\cite{ye2025po3ad}  & - & - & - & - & - & - & - & - & - & - & - & - & - \\ 
 & MC3D-AD\cite{cheng2025mc3d} & 62.8 & 91.0 & 81.9 & 64.0 & 94.2 & 82.2 & 93.2 & 45.8 & 65.9 & 77.8 & 69.0 & {\ul 93.4} & 76.8 \\ 
 & Simple3D\cite{cheng2026towards} & {\ul 88.1} & {\ul 96.2} & \textbf{99.2} & {\ul 86.1} & {\ul 99.0} & {\ul 96.6} & \textbf{96.2} & {\ul 97.3} & \textbf{94.2} & 71.6 & {\ul 85.8} & \textbf{96.8} & {\ul 92.3} \\ 
\multirow{-6}{*}{\rotatebox{90}{P-AUROC}} & \cellcolor[HTML]{D9D9D9}Ours
& \cellcolor[HTML]{D9D9D9}\textbf{91.0}
& \cellcolor[HTML]{D9D9D9}\textbf{98.9}
& \cellcolor[HTML]{D9D9D9}{\ul 95.4}
& \cellcolor[HTML]{D9D9D9}\textbf{87.0}
& \cellcolor[HTML]{D9D9D9}\textbf{99.4}
& \cellcolor[HTML]{D9D9D9}\textbf{96.7}
& \cellcolor[HTML]{D9D9D9}{\ul 93.9}
& \cellcolor[HTML]{D9D9D9}\textbf{98.1}
& \cellcolor[HTML]{D9D9D9}{\ul 89.6}
& \cellcolor[HTML]{D9D9D9}\textbf{88.4}
& \cellcolor[HTML]{D9D9D9}\textbf{87.2}
& \cellcolor[HTML]{D9D9D9}93.2
& \cellcolor[HTML]{D9D9D9}\textbf{93.2} \\ \hline
\end{tabular}}
\end{table}

\subsection{Experimental results on the Real3D-AD dataset}
In this section, we present quantitative and qualitative comparisons with previous methods on the Real3D-AD dataset.
As shown in \cref{tab:real3d-ad}, the proposed method achieves the best performance in terms of both O-AUROC and P-AUROC.
The improvement in O-AUROC confirms that the proposed M2P module effectively suppresses excessive anomaly scores in normal regions by preserving representative prototypes of the normal distribution.
Furthermore, the highest P-AUROC demonstrates the effectiveness of the BSR strategy in mitigating frequent FP along structural boundaries, which are commonly observed in existing memory bank–based approaches.
In terms of P-AUROC, our method achieves either the best or the second-best performance across most object categories.
Notably, clear improvements are observed for classes such as Candybar, Chicken, Duck, and Starfish, which involve intricate geometries and thin structures that typically lead to high false-positive rates near structural boundaries.

\begin{figure}[t]
  \centering
  \includegraphics[width=0.7\linewidth]{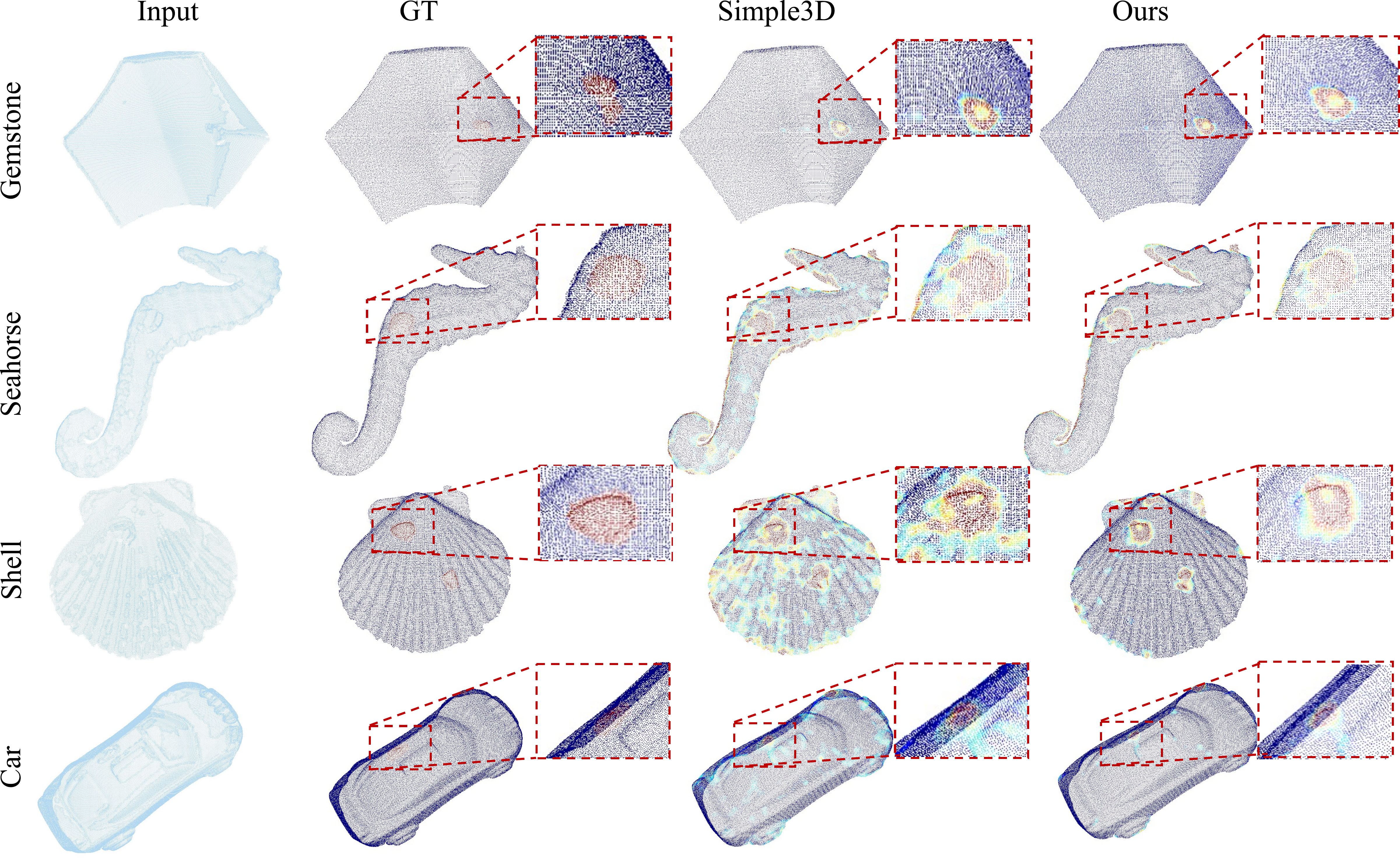}
    \caption{Qualitative comparison of anomaly localization on Real3D-AD. Each row shows the input, GT, and predicted anomaly maps.}
  \label{fig:seyoung3}
\end{figure}
\cref{fig:seyoung3} presents qualitative comparisons of anomaly maps generated by the proposed method and representative existing methods on Real3D-AD.
For the Seahorse class, existing methods exhibit pronounced responses along structural boundaries, resulting in substantial boundary-induced FP. In contrast, the proposed method produces a significantly cleaner anomaly map with markedly reduced boundary artifacts, highlighting the effectiveness of the BSR strategy in refining anomaly scores around complex geometric structures.
In the Shell class, existing approaches suffer from widespread FP both within normal regions and near boundaries. Our method, however, accurately localizes only the true anomalous regions while suppressing false responses elsewhere, demonstrating the effectiveness of the M2P module in more faithfully modeling the normal distribution.

\begin{table}[t]
\caption{Comparison of quantitative results on the Anomaly-ShapeNet. The best performance is in \textbf {bold}, and the second best is {\ul underlined}.}
\label{tab:shapenet}
\setlength{\tabcolsep}{3.5pt} 
\resizebox{\linewidth}{!}{%
\begin{tabular}{l|l|ccccccccccccccccccccc}

\hline
 & Method & Ashtray0 & Bag0 & Bottle0 & Bottle1 & Bottle3 & Bowl0 & Bowl1 & Bowl2 & Bowl3 & Bowl4 & Bowl5 & Bucket0 & Bucket1 & Cap0 & Cap3 & Cap4 & Cap5 & Cup0 & Cup1 & Eraser0 & Headset0 \\ \hline
& IMRNet\cite{li2024towards} & 67.1 & 66.0 & 55.2 & 70.0 & 64.0 & 68.1 & 70.2 & 68.5 & 59.9 & 67.6 & 71.0 & 58.0 & 77.1 & 73.7 & 77.5 & 65.2 & 65.2 & 64.3 & 75.7 & 54.8 & 72.0 \\
 & ISMP\cite{liang2025look} & - & - & - & - & - & - & - & - & - & - & - & - & - & - & - & - & - & - & - & - & - \\
 & PO3AD\cite{ye2025po3ad}  & \textbf{100.0} & 83.3 & {\ul 90.0} & 93.3 & {\ul 92.6} & 92.2 & 82.9 & {\ul 83.3} & 88.1 & \textbf{98.1} & 84.9 & 85.3 & 78.7 & {\ul 87.7} & 85.9 & 79.2 & 67.0 & 87.1 & 83.3 & 99.5 & 80.8 \\
 & MC3D-AD\cite{cheng2025mc3d}  & 96.2 & 80.5 & 79.5 & 70.9 & 75.6 & 93.0 & \textbf{97.8} & 71.9 & 88.5 & 91.1 & 75.4 & {\ul 89.8} & 78.4 & 79.3 & 70.1 & 83.5 & 76.1 & 74.3 & {\ul 95.2} & 77.6 & 86.2 \\
 & Simple3D\cite{cheng2026towards}   & {\ul 99.5} & \textbf{88.1} & \textbf{97.6} & {\ul 95.1} & \textbf{100.0} & \textbf{100.0} & 83.0 & 71.1 & {\ul 91.1} & 73.0 & {\ul 86.3} &  \textbf{95.9} & {\ul 79.0} & 85.2 & {\ul 86.7} & {\ul 91.2} & {\ul 80.4} & \textbf{100.0} & 82.4 & \textbf{100.0} & {\ul 98.2} \\
 & \cellcolor[HTML]{D9D9D9}Ours & \cellcolor[HTML]{D9D9D9}\textbf{100.0} & \cellcolor[HTML]{D9D9D9}{\ul 83.8} & \cellcolor[HTML]{D9D9D9}79.0 & \cellcolor[HTML]{D9D9D9}\textbf{99.3} & \cellcolor[HTML]{D9D9D9}91.4 & \cellcolor[HTML]{D9D9D9}{\ul 99.6} & \cellcolor[HTML]{D9D9D9}{\ul 92.2} & \cellcolor[HTML]{D9D9D9}\textbf{94.1} & \cellcolor[HTML]{D9D9D9}\textbf{97.8} & \cellcolor[HTML]{D9D9D9}{\ul 93.7} & \cellcolor[HTML]{D9D9D9}\textbf{97.2} & \cellcolor[HTML]{D9D9D9}89.5 & \cellcolor[HTML]{D9D9D9}\textbf{81.0} & \cellcolor[HTML]{D9D9D9}\textbf{88.1} & \cellcolor[HTML]{D9D9D9}\textbf{87.7} & \cellcolor[HTML]{D9D9D9}\textbf{94.4} & \cellcolor[HTML]{D9D9D9}\textbf{82.1} & \cellcolor[HTML]{D9D9D9}{\ul 99.0} & \cellcolor[HTML]{D9D9D9}\textbf{99.5} & \cellcolor[HTML]{D9D9D9}{\ul 99.0} & \cellcolor[HTML]{D9D9D9}\textbf{100.0} \\ \cline{2-23} 
 &  & Headset1 & Helmet0 & Helmet1 & Helmet2 & Helmet3 & Jar0 & Microphone0 & Shelf0 & Tap0 & Tap1 & Vase0 & Vase1 & Vase2 & Vase3 & Vase4 & Vase5 & Vase7 & Vase8 & Vase9 & \multicolumn{2}{|c}{Mean} \\ \cline{2-23} 
 & IMRNet\cite{li2024towards}  & 67.6 & 59.7 & 60.0 & 64.1 & 57.3 & 78.0 & 75.5 & 60.3 & 67.6 & {\ul 69.6} & 53.3 & 75.7 & 61.4 & 70.0 & 52.4 & 67.6 & 63.5 & 63.0 & 59.4 & \multicolumn{2}{|c}{66.1} \\
 & ISMP\cite{liang2025look}  & - & - & - & - & - & - & - & - & - & - & - & - & - & - & - & - & - & - & - & \multicolumn{2}{|c}{75.7} \\
 & PO3AD\cite{ye2025po3ad}  & {\ul 92.3} & {\ul 76.2} & {\ul 96.1} & \textbf{86.9} & 75.4 & 86.6 & 77.6 & 57.3 & 74.5 & 68.1 & 85.8 & 74.2 & \textbf{95.2} & \textbf{82.1} & 67.5 & 85.2 & \textbf{96.6} & 73.9 & \textbf{83.0} & \multicolumn{2}{|c}{83.9} \\
 & MC3D-AD\cite{cheng2025mc3d}  & 88.6 & 67.2 & \textbf{100.0} & 60.9 & \textbf{97.9} & \textbf{97.1} & {\ul 91.9} & \textbf{84.1} & \textbf{94.5} & \textbf{97.0} & 82.1 & {\ul 85.7} & {\ul 92.9} & 76.1 & \textbf{87.6} & 85.2 & 93.8 & 67.0 & 73.6 & \multicolumn{2}{|c}{84.2} \\
 & Simple3D\cite{cheng2026towards}  & \textbf{95.7} & 69.0 & 71.9 & 75.4 & 65.5 & {\ul 90.5} & \textbf{100.0} & 74.8 & 70.3 & 59.6 & {\ul 95.4} & 82.4 & 87.1 & {\ul 81.2} & {\ul 86.4} & {\ul 96.2} & 89.5 & {\ul 85.5} & {\ul 81.5} & \multicolumn{2}{|c}{{\ul 86.0}} \\
  \multirow{-14}{*}{\rotatebox{90}{O-AUROC}}
  & \cellcolor[HTML]{D9D9D9}Ours & \cellcolor[HTML]{D9D9D9}65.7 & \cellcolor[HTML]{D9D9D9}\textbf{86.7} & \cellcolor[HTML]{D9D9D9}81.4 & \cellcolor[HTML]{D9D9D9}{\ul 83.8} & \cellcolor[HTML]{D9D9D9}{\ul 78.5} & \cellcolor[HTML]{D9D9D9}82.4 & \cellcolor[HTML]{D9D9D9}\textbf{100.0} & \cellcolor[HTML]{D9D9D9}{\ul 80.3} & \cellcolor[HTML]{D9D9D9}{\ul 79.7} & \cellcolor[HTML]{D9D9D9}65.6 & \cellcolor[HTML]{D9D9D9}\textbf{99.6} & \cellcolor[HTML]{D9D9D9}\textbf{96.2} & \cellcolor[HTML]{D9D9D9}90.0 & \cellcolor[HTML]{D9D9D9}{\ul 81.2} & \cellcolor[HTML]{D9D9D9}83.3 & \cellcolor[HTML]{D9D9D9}\textbf{99.0} & \cellcolor[HTML]{D9D9D9}{\ul 95.7} & \cellcolor[HTML]{D9D9D9}\textbf{87.0} & \cellcolor[HTML]{D9D9D9}79.7 & \multicolumn{2}{|c}{\cellcolor[HTML]{D9D9D9}\textbf{89.1}} \\ \hline

 & Method & Ashtray0 & Bag0 & Bottle0 & Bottle1 & Bottle3 & Bowl0 & Bowl1 & Bowl2 & Bowl3 & Bowl4 & Bowl5 & Bucket0 & Bucket1 & Cap0 & Cap3 & Cap4 & Cap5 & Cup0 & Cup1 & Eraser0 & Headset0 \\ \hline
& IMRNet\cite{li2024towards} & - & - & - & - & - & - & - & - & - & - & - & - & - & - & - & - & - & - & - & - & - \\
 & ISMP\cite{liang2025look}  & 60.3 & 74.7 & 77.0 & 56.8 & 77.5 & 85.1 & 54.6 & 73.6 & 77.3 & 74.0 & 53.4 & 52.4 & 67.2 & 86.5 & 73.4 & 75.3 & 67.8 & 86.9 & 60.0 & 70.6 & 58.0 \\
 & PO3AD\cite{ye2025po3ad} & \textbf{96.2} & {\ul 94.9} & 91.2 & \textbf{84.4} & \textbf{88.0} & 97.8 & 91.4 & 91.8 & 93.5 & {\ul 96.7} & 94.1 & \textbf{75.5} & 89.9 & 95.7 & 94.8 & 94.0 & 86.4 & 90.9 & 93.2 & \textbf{97.4} & 82.3 \\
 & MC3D-AD\cite{cheng2025mc3d}  & - & - & - & - & - & - & - & - & - & - & - & - & - & - & - & - & - & - & - & - & - \\
 & Simple3D\cite{cheng2026towards}  & 92.0 & \textbf{95.4} & \textbf{97.4} & 72.8 & {\ul 83.8} & {\ul 98.8} & {\ul 95.1} & {\ul 93.3} & {\ul 99.3} & 92.9 & \textbf{97.9} & {\ul 72.5} & {\ul 92.1} & \textbf{98.8} & {\ul 96.4} & \textbf{97.9} & {\ul 96.4} & {\ul 97.9} & {\ul 93.7} & {\ul 97.0} & {\ul 93.6} \\
 & \cellcolor[HTML]{D9D9D9}Ours & \cellcolor[HTML]{D9D9D9}{\ul 94.6} & \cellcolor[HTML]{D9D9D9}94.6 & \cellcolor[HTML]{D9D9D9}{\ul 94.2} & \cellcolor[HTML]{D9D9D9}{\ul 74.0} & \cellcolor[HTML]{D9D9D9}80.9 & \cellcolor[HTML]{D9D9D9}\textbf{99.4} & \cellcolor[HTML]{D9D9D9}\textbf{96.8} & \cellcolor[HTML]{D9D9D9}\textbf{98.1} & \cellcolor[HTML]{D9D9D9}\textbf{99.5} & \cellcolor[HTML]{D9D9D9}\textbf{99.1} & \cellcolor[HTML]{D9D9D9}{\ul 97.8} & \cellcolor[HTML]{D9D9D9}71.1 & \cellcolor[HTML]{D9D9D9}\textbf{95.7} & \cellcolor[HTML]{D9D9D9}{\ul 98.2} & \cellcolor[HTML]{D9D9D9}\textbf{97.8} & \cellcolor[HTML]{D9D9D9}{\ul 97.4} & \cellcolor[HTML]{D9D9D9}\textbf{97.6} & \cellcolor[HTML]{D9D9D9}\textbf{98.8} & \cellcolor[HTML]{D9D9D9}\textbf{95.1} & \cellcolor[HTML]{D9D9D9}96.8 & \cellcolor[HTML]{D9D9D9}\textbf{97.2} \\ \cline{2-23} 
 &  & Headset1 & Helmet0 & Helmet1 & Helmet2 & Helmet3 & Jar0 & Microphone0 & Shelf0 & Tap0 & Tap1 & Vase0 & Vase1 & Vase2 & Vase3 & Vase4 & Vase5 & Vase7 & Vase8 & Vase9 & \multicolumn{2}{|c}{Mean} \\ \cline{2-23} 
 & IMRNet\cite{li2024towards}  & - & - & - & - & - & - & - & - & - & - & - & - & - & - & - & - & - & - & - & \multicolumn{2}{|c}{-} \\
 & ISMP\cite{liang2025look}  & 70.2 & 68.3 & 62.2 & 84.4 & 72.2 & 82.3 & 66.1 & 68.7 & 52.2 & 55.2 & 66.1 & 84.3 & 73.3 & 76.2 & 54.5 & 47.2 & 70.1 & 85.1 & 61.5 & \multicolumn{2}{|c}{69.1} \\
 & PO3AD\cite{ye2025po3ad}  & 90.7 & 87.8 & {\ul 94.8} & 93.2 & 84.6 & 87.1 & 81.0 & 66.3 & 78.3 & 69.2 & \textbf{95.5} & {\ul 88.2} & 97.8 & 88.4 & 90.2 & 93.7 & 98.2 & 95.0 & \textbf{95.2} & \multicolumn{2}{|c}{89.8} \\
 & MC3D-AD\cite{cheng2025mc3d} & - & - & - & - & - & - & - & - & - & - & - & - & - & - & - & - & - & - & - & \multicolumn{2}{|c}{-} \\
 & Simple3D\cite{cheng2026towards}  & \textbf{95.8} & {\ul 90.2} & 90.2 & {\ul 94.7} & {\ul 92.8} & \textbf{97.8} & {\ul 96.4} & {\ul 90.1} & {\ul 85.7} & {\ul 81.1} & {\ul 93.4} & 80.7 & {\ul 98.6} & {\ul 91.0} & {\ul 98.5} & {\ul 96.6} & {\ul 99.0} & {\ul 97.5} & 91.3 & \multicolumn{2}{|c}{{\ul 92.9}} \\
  \multirow{-14}{*}{\rotatebox{90}{P-AUROC}}
 & \cellcolor[HTML]{D9D9D9}Ours & \cellcolor[HTML]{D9D9D9}{\ul 92.5} & \cellcolor[HTML]{D9D9D9}\textbf{92.6} & \cellcolor[HTML]{D9D9D9}\textbf{96.0} & \cellcolor[HTML]{D9D9D9}\textbf{95.2} & \cellcolor[HTML]{D9D9D9}\textbf{97.1} & \cellcolor[HTML]{D9D9D9}{\ul 97.5} & \cellcolor[HTML]{D9D9D9}\textbf{97.0} & \cellcolor[HTML]{D9D9D9}\textbf{91.6} & \cellcolor[HTML]{D9D9D9}\textbf{89.7} & \cellcolor[HTML]{D9D9D9}\textbf{83.9} & \cellcolor[HTML]{D9D9D9}{\ul 93.4} & \cellcolor[HTML]{D9D9D9}\textbf{90.5} & \cellcolor[HTML]{D9D9D9}\textbf{98.8} & \cellcolor[HTML]{D9D9D9}\textbf{94.9} & \cellcolor[HTML]{D9D9D9}\textbf{98.7} & \cellcolor[HTML]{D9D9D9}\textbf{98.3} & \cellcolor[HTML]{D9D9D9}\textbf{99.3} & \cellcolor[HTML]{D9D9D9}\textbf{98.8} & \cellcolor[HTML]{D9D9D9}{\ul 92.7} & \multicolumn{2}{|c}{\cellcolor[HTML]{D9D9D9}\textbf{94.3}} \\ \cline{1-23} 
\end{tabular}}
\end{table}

\begin{figure}[!b]
  \centering
  \includegraphics[width=0.7\linewidth]{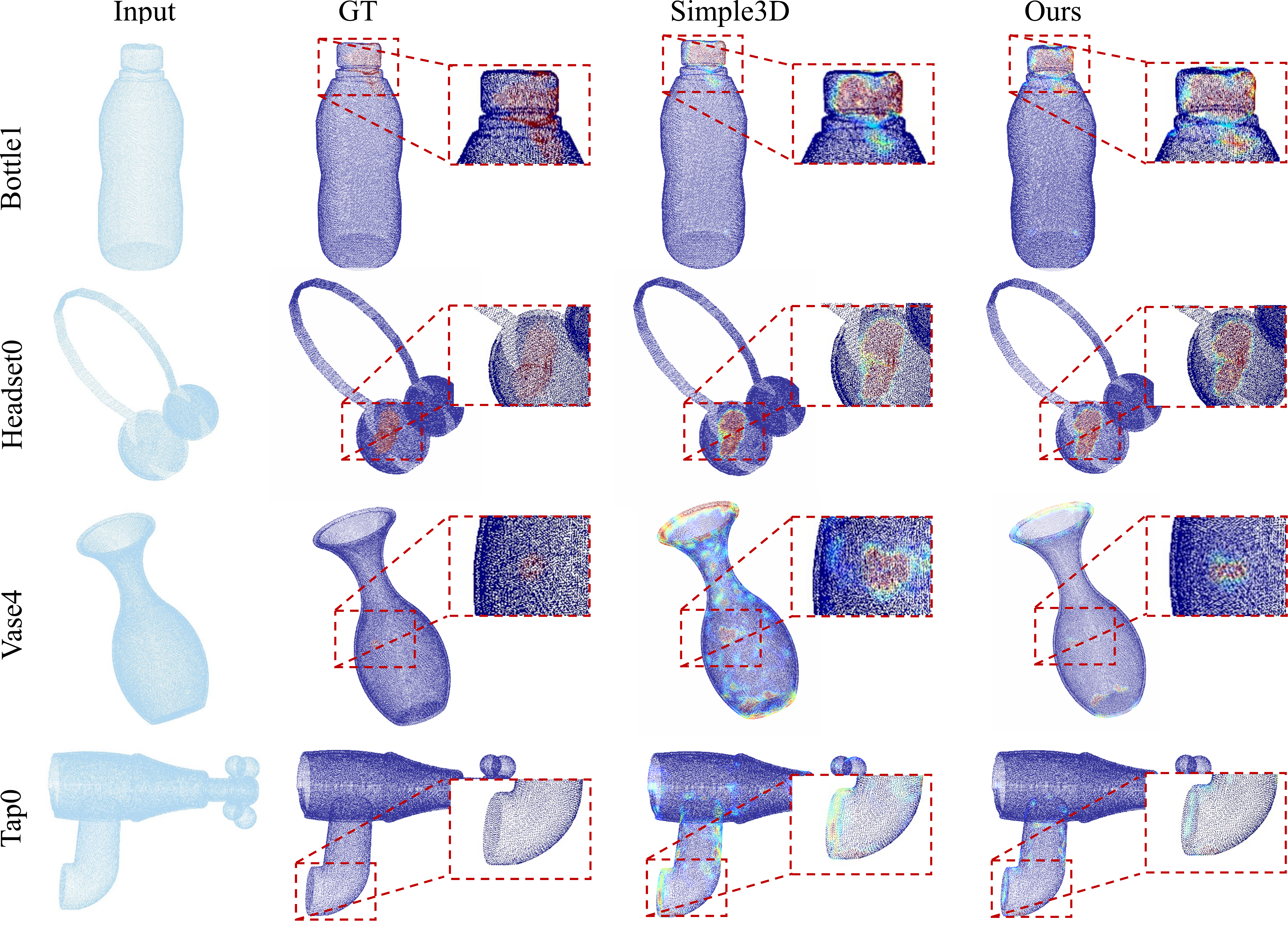}
    \caption{Qualitative comparison of anomaly localization on Anomaly-ShapeNet. Each row shows the input, GT, and predicted anomaly maps.}
  \label{fig:seyoung4}
\end{figure}

\subsection{Experimental results on the Anomaly-ShapeNet dataset}
We present quantitative and qualitative comparisons with previous methods on the Anomaly-ShapeNet dataset.
As shown in \cref{tab:shapenet}, our method achieves state-of-the-art performance with an O-AUROC of 89.1\% and a P-AUROC of 94.3\%.
Notably, the proposed approach attains the best P-AUROC in most object categories, demonstrating strong localization capability and robust generalization across diverse shapes.
To evaluate the proposed method qualitatively on the Anomaly-ShapeNet dataset, we present comparisons of anomaly maps generated by the proposed method and representative existing methods in \cref{fig:seyoung4}.
In particular, for the Vase4 class, the proposed M2P module significantly reduces FP in normal regions that frequently occur in previous methods, resulting in cleaner anomaly localization.
Moreover, boundary-induced FP are effectively suppressed by the BSR, leading to substantially cleaner anomaly maps along object boundaries.
For the Bottle1 class, where the proposed method shows relatively lower quantitative performance compared to existing baselines, the predicted anomaly maps remain visually consistent with the ground-truth annotations, accurately localizing anomalous regions.

\begin{table}[t]
\caption{Comparison of quantitative results on MulSen-AD. The best performance is in \textbf {bold}, and the second best is {\ul underlined}.}
\label{tab:mulsen-ad}
\centering
\setlength{\tabcolsep}{3.5pt} 
\resizebox{\linewidth}{!}{%
\begin{tabular}{c|l|ccccccccccccccc|c}
\hline
& Method & Capsule & Cotton & Cube & \begin{tabular}[c]{@{}c@{}}Spring\\[-0.7mm]pad\end{tabular}   & Screw & Screen & Piggy & Nut & \begin{tabular}[c]{@{}c@{}}Flat\\[-0.7mm]pad\end{tabular} & \begin{tabular}[c]{@{}c@{}}Plastic\\[-0.7mm]cylinder\end{tabular}   & Zipper & \begin{tabular}[c]{@{}c@{}}Button\\[-0.7mm]cell\end{tabular}   & \begin{tabular}[c]{@{}c@{}}Tooth\\[-0.7mm]brush\end{tabular} & \begin{tabular}[c]{@{}c@{}}Solar\\[-0.7mm]panel\end{tabular}  & Light & Mean \\ \hline
 & Reg3D-AD\cite{liu2023real3d}  & 91.2 & 43.0 & 56.9 & {\ul 95.1} & \textbf{97.2} & 64.1 & 86.6 & 79.7 & 90.8 & 76.5 & 47.0 & 78.2 & 81.2 & {\ul 66.0} & 89.7 & 74.9 \\
 & IMRNet\cite{li2024towards} & 60.1 & 58.5 & 43.2 & 65.1 & 74.2 & 37.8 & 72.9 & 81.2 & 71.4 & 62.1 & 63.0 & 70.2 & 61.5 & 34.4 & 45.7 & 60.1 \\
 & GLFM\cite{cheng2025boosting}  & {\ul 96.7} & \textbf{81.2} & 75.6 & \textbf{100.0} & 63.6 & {\ul 86.6} & 73.2 & 94.0 & {\ul 94.6} & 81.5 & 81.3 & 56.7 & \textbf{84.9} & 40.9 & 66.1 & 78.5 \\
 & Simple3D\cite{cheng2026towards}  & \textbf{98.1} & {\ul 78.8} & {\ul 82.7} & \textbf{100.0} & 86.4 & 73.8 & {\ul 89.6} & {\ul 96.9} & \textbf{100.0} & \textbf{99.4} & {\ul 88.5} & {\ul 85.1} & {\ul 82.9} & 64.4 & \textbf{96.8} & {\ul 88.2} \\ 
\multirow{-5}{*}{\rotatebox{90}{O-AUROC}}  & \cellcolor[HTML]{D9D9D9}Ours & \cellcolor[HTML]{D9D9D9}96.2 & \cellcolor[HTML]{D9D9D9}78.0 & \cellcolor[HTML]{D9D9D9}\textbf{92.0} & \cellcolor[HTML]{D9D9D9}\textbf{100.0} & \cellcolor[HTML]{D9D9D9}{\ul 90.5} & \cellcolor[HTML]{D9D9D9}\textbf{90.9} & \cellcolor[HTML]{D9D9D9}\textbf{97.5} & \cellcolor[HTML]{D9D9D9}\textbf{97.1} & \cellcolor[HTML]{D9D9D9}\textbf{100.0} & \cellcolor[HTML]{D9D9D9}{\ul 98.0} & \cellcolor[HTML]{D9D9D9}\textbf{93.7} & \cellcolor[HTML]{D9D9D9}\textbf{90.3} & \cellcolor[HTML]{D9D9D9}81.6 & \cellcolor[HTML]{D9D9D9}\textbf{71.1} & \cellcolor[HTML]{D9D9D9}{\ul 96.1} & \cellcolor[HTML]{D9D9D9}\textbf{91.5} \\ \hline
 & Reg3D-AD\cite{liu2023real3d}  & 87.7 & 52.1 & 62.6 & 80.2 & 54.0 & 46.6 & 63.5 & 80.7 & 69.2 & 67.0 & 53.6 & 70.6 & 47.2 & 60.9 & {\ul 65.1} & 64.1 \\
 & IMRNet\cite{li2024towards} & 42.3 & 50.7 & 56.6 & 40.1 & 45.6 & 35.2 & 51.2 & 36.9 & 54.2 & 41.2 & 49.6 & 48.5 & 51.9 & 53.3 & 41.5 & 46.7 \\
 & GLFM\cite{cheng2025boosting} & 93.0 & {\ul 67.9} & 68.0 & 70.5 & 60.9 & 50.8 & 77.8 & 95.7 & 77.2 & 67.8 & {\ul 57.4} & 43.3 & 57.8 & 57.5 & 52.6 & 66.5 \\
 & Simple3D\cite{cheng2026towards}  & \textbf{97.3} & 60.2 & {\ul 73.2} & {\ul 81.1} & \textbf{74.7} & {\ul 52.9} & {\ul 91.6} & \textbf{97.2} & {\ul 82.0} & {\ul 96.4} & 53.0 & {\ul 80.1} & \textbf{73.3} & {\ul 94.9} & \textbf{95.9} & {\ul 80.3} \\
\multirow{-5}{*}{\rotatebox{90}{P-AUROC}}  & \cellcolor[HTML]{D9D9D9}Ours & \cellcolor[HTML]{D9D9D9}{\ul 95.8} & \cellcolor[HTML]{D9D9D9}\textbf{71.7} & \cellcolor[HTML]{D9D9D9}\textbf{94.4} & \cellcolor[HTML]{D9D9D9}\textbf{82.4} & \cellcolor[HTML]{D9D9D9}{\ul 72.4} & \cellcolor[HTML]{D9D9D9}\textbf{64.4} & \cellcolor[HTML]{D9D9D9}\textbf{95.4} & \cellcolor[HTML]{D9D9D9}{\ul 96.9} & \cellcolor[HTML]{D9D9D9}\textbf{84.5} & \cellcolor[HTML]{D9D9D9}\textbf{97.6} & \cellcolor[HTML]{D9D9D9}\textbf{62.0} & \cellcolor[HTML]{D9D9D9}\textbf{86.6} & \cellcolor[HTML]{D9D9D9}{\ul 70.5} & \cellcolor[HTML]{D9D9D9}\textbf{97.2} & \cellcolor[HTML]{D9D9D9}\textbf{95.9} & \cellcolor[HTML]{D9D9D9}\textbf{84.6} \\ \hline
\end{tabular}}
\end{table}

\begin{figure}[t]
  \centering
  \includegraphics[width=0.7\linewidth]{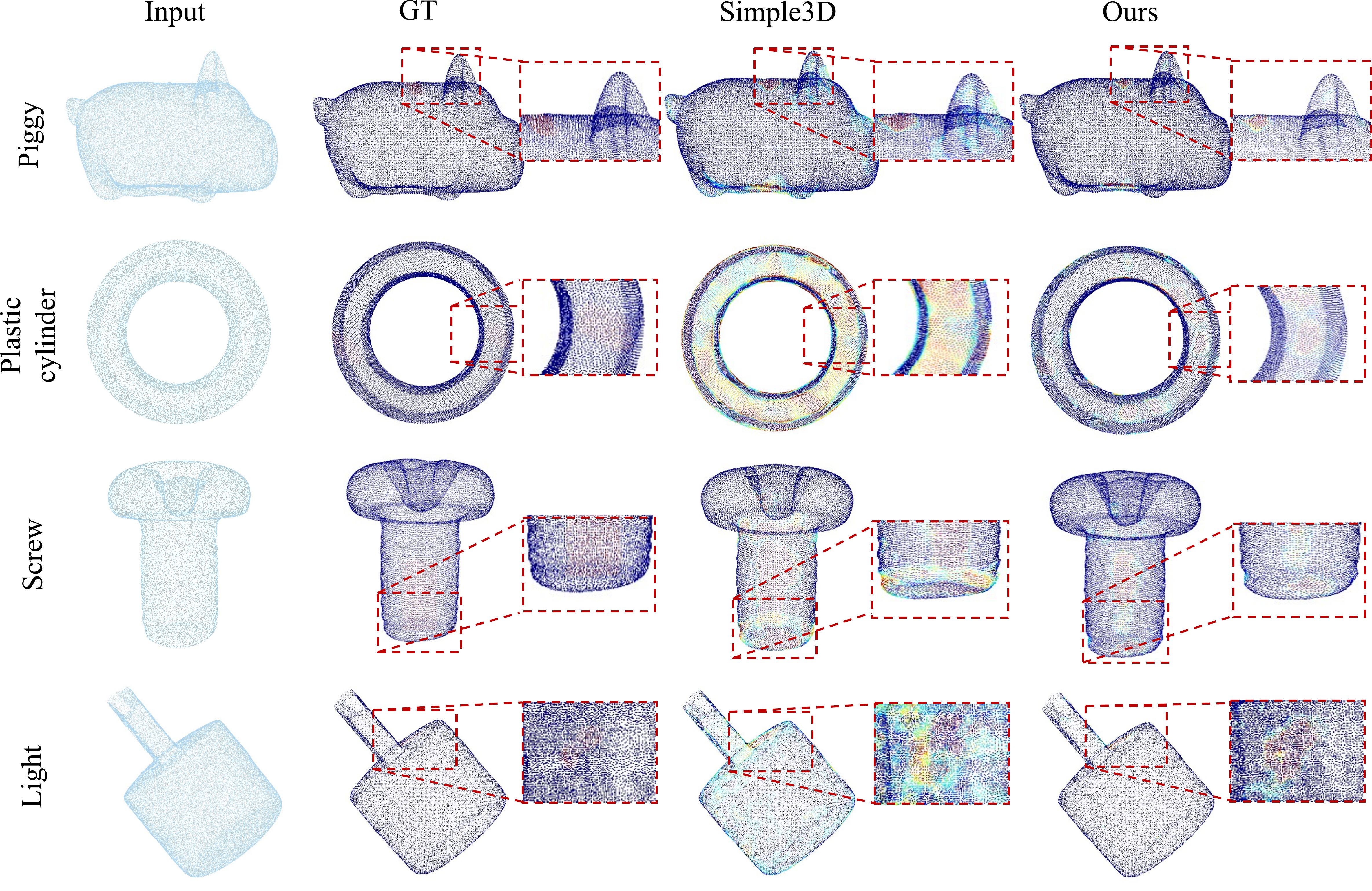}
    \caption{Qualitative comparison of anomaly localization on MulSen-AD. Each row shows the input, GT, and predicted anomaly maps.}
  \label{fig:seyoung5}
\end{figure}

\subsection{Experimental results on the MulSen-AD dataset}
In this section, we present quantitative and qualitative comparisons with previous methods on the MulSen-AD dataset.
\cref{tab:mulsen-ad} presents a performance comparison in terms of O-AUROC and P-AUROC. 
The proposed M2P-AD achieves significant improvements of 3.3\% and 4.3\% over the baseline model in O-AUROC and P-AUROC, respectively. 
Moreover, it attains the best performance in most category classes, demonstrating strong generalization across diverse object types.
In particular, notable improvements are observed in categories with sharp and angular geometries, such as Cube, Screen, and Solar panel, suggesting the effectiveness of the BSR in handling complex structural boundaries.
To evaluate the proposed method qualitatively on the MulSen-AD dataset, we present comparisons of anomaly maps predicted by the proposed method and representative existing methods in \cref{fig:seyoung5}.
Compared to baseline models, the proposed method significantly reduces FP while more precisely localizing anomalous regions.
The results qualitatively validate the effectiveness of the M2P module and BSR across diverse industrial environments and categories.

\section{Ablation study}

\begin{figure}[h]
  \centering
  \includegraphics[width=0.7\linewidth]{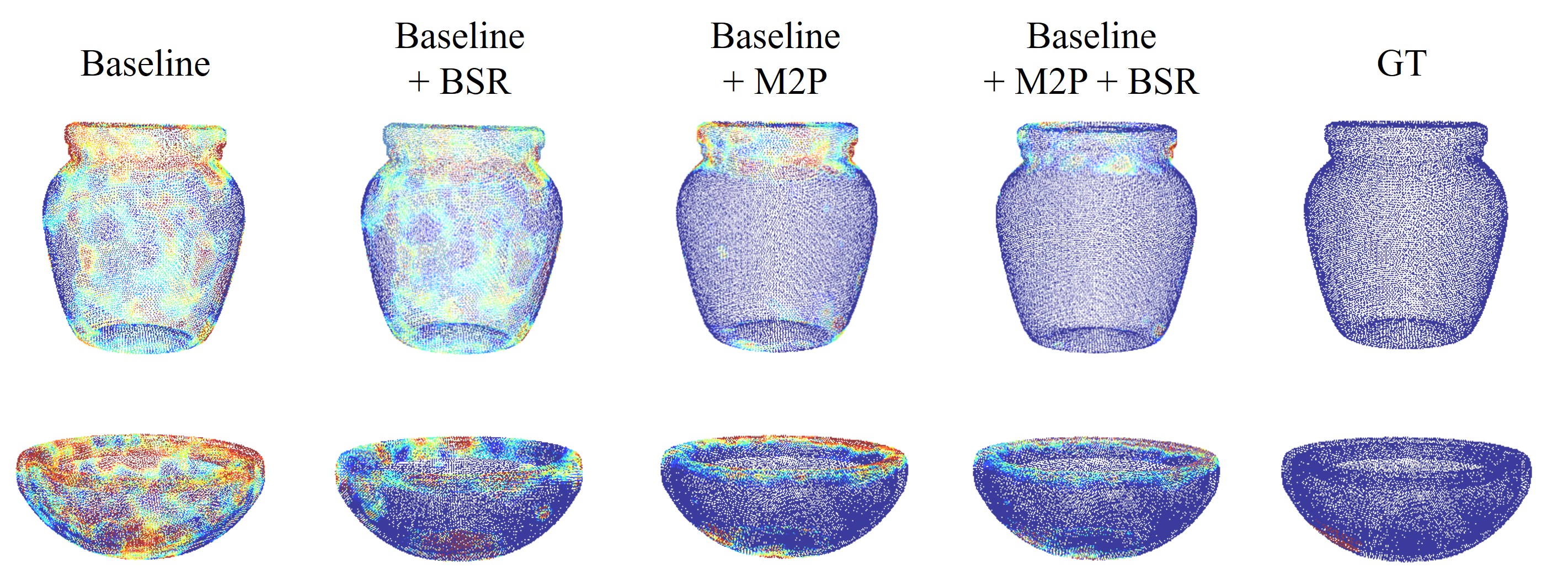}
    \caption{Qualitative visualization of the effectiveness of the M2P module and BSR.}
  \label{fig:seyoung6}
\end{figure}

\noindent\textbf{Effectiveness of M2P and BSR.}
We conduct a comprehensive ablation study on three benchmark datasets, including Real3D-AD, Anomaly-ShapeNet, and MulSen-AD, to evaluate the individual contributions of the proposed M2P module and BSR strategy.
As shown in \cref{tab:ablation_1}, both modules independently improve performance across all datasets and evaluation metrics.
Notably, the combination of M2P and BSR achieves the best overall performance on all three datasets, indicating that the two components integrate effectively.
As shown in \cref{fig:seyoung6}, the anomaly score maps obtained by individually applying M2P and BSR demonstrate the distinct contribution of each component. 
M2P reduces excessive anomaly responses in normal regions, while BSR alleviates FP near object boundaries by considering boundary characteristics, resulting in more accurate and robust localization.
These results demonstrate that both components effectively address the key challenges in 3D anomaly detection, namely excessive anomaly scores in normal regions and FP near object boundaries, leading to substantial and consistent improvements across diverse datasets.
\begin{table}[h]
\caption{Ablation study on the effectiveness of the M2P module and the BSR.}
\label{tab:ablation_1}
\centering
\setlength{\tabcolsep}{4.5pt} 
\resizebox{0.7\linewidth}{!}{%
\begin{tabular}{c|c|cc|cc|cc}
\hline
\multirow{2}{*}{M2P} & \multirow{2}{*}{BSR} & \multicolumn{2}{c|}{Real3D-AD} & \multicolumn{2}{c|}{Anomaly-ShapeNet} & \multicolumn{2}{c}{MulSen-AD} \\ \cline{3-8} 
 &  & O-AUROC & P-AUROC & O-AUROC & P-AUROC & O-AUROC & P-AUROC \\ \hline
\multicolumn{1}{l|}{} & \multicolumn{1}{l|}{} & 80.4 & 92.3 & 86.0 & 92.9 & 88.2 & 80.3 \\
\multicolumn{1}{l|}{} & \checkmark & 81.5 & 92.9 & 88.9 & 93.9 & 90.8 & 84.4 \\
\checkmark & \multicolumn{1}{l|}{} & 82.0 & 93.0 & 88.9 & \textbf{94.3} & 91.2 & 84.4 \\
\checkmark & \checkmark & \textbf{82.4} & \textbf{93.2} & \textbf{89.1} & \textbf{94.3} & \textbf{91.5} & \textbf{84.6} \\ \hline
\end{tabular}}
\end{table}

\begin{table}[t]
\caption{Ablation study on the effect of the hyperparameter $K$ in the M2P module.}
\label{tab:ablation_2}
\centering
\setlength{\tabcolsep}{4.5pt} 
\resizebox{0.65\linewidth}{!}{%
\begin{tabular}{>{\centering\arraybackslash}p{0.6cm}|cc|cc|cc}
\hline
\multirow{2}{*}{$K$} & \multicolumn{2}{c|}{Real3D-AD} & \multicolumn{2}{c|}{Anomaly-ShapeNet} & \multicolumn{2}{c}{MulSen-AD} \\ \cline{2-7} 
 & O-AUROC & P-AUROC & O-AUROC & P-AUROC & O-AUROC & P-AUROC \\ \hline
32 & 82.1 & 93.0 & 88.9 & 93.9 & 91.1 & 84.1 \\
64 & 81.9 & 93.0 & 88.8 & 94.0 & 91.4 & 84.3 \\
128 & \textbf{82.4} & \textbf{93.2} & \textbf{89.1} & \textbf{94.3} & 91.5 & \textbf{84.6} \\
256 & 81.8 & 92.9 & 88.9 & 93.2 & \textbf{91.7} & 84.5 \\ \hline
\end{tabular}}
\end{table}

\begin{table}[!h]
\caption{Ablation study on the effect of the hyperparameter $n$ in BSR.}
\label{tab:ablation_3}
\centering
\setlength{\tabcolsep}{4.5pt} 
\resizebox{0.65\linewidth}{!}{%
\begin{tabular}{>{\centering\arraybackslash}p{0.6cm}|cc|cc|cc}
\hline
\multirow{2}{*}{$n$} & \multicolumn{2}{c|}{Real3D-AD} & \multicolumn{2}{c|}{Anomaly-ShapeNet} & \multicolumn{2}{c}{MulSen-AD} \\  \cline{2-7} 
 & O-AUROC & P-AUROC & O-AUROC & P-AUROC & O-AUROC & P-AUROC \\ \hline
5 & 82.0 & 93.0 & 88.8 & 94.1 & 90.6 & 84.5 \\
10 & 82.0 & 93.1 & 88.7 & 94.2 & 90.6 & 84.4 \\
20 & \textbf{82.4} & \textbf{93.2} & \textbf{89.1} & \textbf{94.3} & \textbf{91.5} & \textbf{84.6} \\
30 & 82.1 & 93.0 & 88.8 & \textbf{94.3} & 90.8 & 84.1 \\ \hline
\end{tabular}}
\end{table}

\noindent\textbf{Hyperparameter analysis of the M2P module and BSR.}
We conduct ablation studies by varying the number of clusters $K$ in the M2P module and the top-$n\%$ selection ratio in the BSR strategy.
As shown in \cref{tab:ablation_2}, the best overall performance across most evaluation metrics is achieved when $K=128$.
When $K$ is too small, heterogeneous normal regions are represented by a limited set of prototypes, which blurs distinctive characteristics and introduces ambiguity in similarity-based matching. 
In contrast, an excessively large $K$ over-fragments coherent structural patterns across numerous prototypes, thereby weakening their representativeness. 
When $K=128$, the M2P module achieves a balance between compactness and discriminability in modeling the normal distribution. 

As shown in \cref{tab:ablation_3}, the best overall performance is achieved when $n=20$.
For the Real3D-AD and Anomaly-ShapeNet datasets, performance variations across different values of $n$ are relatively small. 
This can be attributed to the limited number of training samples per class, as each class contains only four normal samples, which constrains the diversity of boundary features stored in the memory bank and leads to minimal performance fluctuations across different settings.
In contrast, the MulSen-AD dataset shows more noticeable changes, suggesting that the optimal selection ratio has a greater impact when richer training data are available.
Overall, these results indicate that selecting an appropriate top-$n\%$ ratio is important for stable and effective boundary-aware score refinement. \newline

\noindent\textbf{Analysis of Loss Functions.}
To assess their effects on prototype learning, we compare different loss configurations in the M2P module.
As presented in \cref{tab:ablation_4}, combining the L2 and cosine losses consistently yields the best performance across all evaluation metrics.
Since each prototype compresses multiple normal patch embeddings into a single representation, minimizing only the Euclidean distance may overemphasize variations in feature magnitude and lead to a biased prototype representation.
Consequently, the learned prototype can become locally compact but may insufficiently preserve the common structural tendency shared by its assigned normal patch embeddings.
The cosine loss complements this objective by reducing sensitivity to feature magnitude and encouraging the prototype to align with the dominant feature direction of the assigned normal patches.
Therefore, the combined objective enables each prototype to better represent the underlying normal structure.\newline

\begin{table}[t]
\caption{Ablation study on the effect of loss configurations in the M2P module.}
\label{tab:ablation_4}
\centering
\setlength{\tabcolsep}{4.5pt} 
\resizebox{0.7\linewidth}{!}{%
\begin{tabular}{c|c|c|cc|cc|cc}
\hline
\multicolumn{3}{c|}{Loss} & \multicolumn{2}{c|}{Real3D-AD} & \multicolumn{2}{c|}{Anomaly-ShapeNet} & \multicolumn{2}{c}{MulSen-AD} \\ \hline
L1 & L2 & Cosine & \multicolumn{1}{c}{O-AUROC} & \multicolumn{1}{c|}{P-AUROC} & \multicolumn{1}{c}{O-AUROC} & \multicolumn{1}{c|}{P-AUROC} & \multicolumn{1}{c}{O-AUROC} & \multicolumn{1}{c}{P-AUROC} \\ \hline
\checkmark &  &  & 80.9 & 91.9 & 86.9 & 92.2 & 88.8 & 83.4 \\
 & \checkmark &  & 81.5 & 92.4 & 88.7 & 94.0 & 91.1 & 84.0 \\
 &  & \checkmark & 81.5 & 92.7 & 88.4 & 92.9 & 90.5 & 82.2 \\
 & \checkmark & \checkmark & \textbf{82.4} & \textbf{93.2} & \textbf{89.1} & \textbf{94.3} & \textbf{91.5} & \textbf{84.6} \\ \hline
\end{tabular}}
\end{table}

\begin{table}[t]
\caption{Ablation study on feature extractors.}
\label{tab:ablation_5}
\centering
\setlength{\tabcolsep}{4.5pt} 
\resizebox{0.67\linewidth}{!}{%
\begin{tabular}{l|cc|cc|cc}
\hline
Feature & \multicolumn{2}{c|}{Real3D-AD} & \multicolumn{2}{c|}{Anomaly-ShapeNet} & \multicolumn{2}{c}{MulSen-AD} \\ \cline{2-7} 
extractor & O-AUROC & P-AUROC & O-AUROC & P-AUROC & O-AUROC & P-AUROC \\ \hline
PointMAE & 81.8 & 92.2 & 88.2 & 93.9 & 90.3 & 83.9 \\
PointBERT & 80.6 & 91.1 & 88.9 & 92.8 & 89.6 & 82.7 \\ 
FPFH & \textbf{82.4} & \textbf{93.2} & \textbf{89.1} & \textbf{94.3} & \textbf{91.5} & \textbf{84.6} \\ \hline
\end{tabular}}
\end{table}

\noindent\textbf{Analysis of feature extractors.}
To assess the generality of the proposed method, performance is evaluated using various feature extractors.
As presented in \cref{tab:ablation_5}, the best performance is obtained when FPFH is employed, while competitive results are also achieved with learning-based feature extractors.
These results demonstrate the generalization capability of the proposed method across diverse feature representations.

\section{Conclusion}
In this paper, we proposed a novel 3D anomaly detection framework, M2P-AD, designed to address the challenge of excessive anomaly scores in normal regions and frequent FP near object boundaries.
The proposed method introduces an M2P module that learns representative prototypes to model the normal distribution while preserving structurally important normal-region information.
In addition, to mitigate FP near object boundaries, a BE module is introduced to extract boundary features, which are leveraged by the BSR strategy for score refinement.
Experimental results on multiple benchmark datasets demonstrate that the proposed method consistently outperforms existing approaches in quantitative performance and achieves state-of-the-art results. 
Furthermore, qualitative results demonstrate that the proposed method localizes anomalies more accurately and robustly while effectively reducing excessive anomaly responses in normal regions and suppressing boundary-induced FP.
Overall, the proposed framework provides an effective solution for reliable 3D anomaly detection.

%
%
\bibliographystyle{splncs04}
\bibliography{main}

@String(AAAI  = {AAAI})

@String(ICPR  = {Int. Conf. Pattern Recog.})

@String(ICPR  = {ICPR})

@article{liu2023real3d,
  title={Real3d-ad: A dataset of point cloud anomaly detection},
  author={Liu, Jiaqi and Xie, Guoyang and Chen, Ruitao and Li, Xinpeng and Wang, Jinbao and Liu, Yong and Wang, Chengjie and Zheng, Feng},
  journal={Advances in Neural Information Processing Systems},
  volume={36},
  pages={30402--30415},
  year={2023}
}

@inproceedings{chen2023easynet,
  title={Easynet: An easy network for 3d industrial anomaly detection},
  author={Chen, Ruitao and Xie, Guoyang and Liu, Jiaqi and Wang, Jinbao and Luo, Ziqi and Wang, Jinfan and Zheng, Feng},
  booktitle={Proceedings of the 31st ACM International Conference on Multimedia},
  pages={7038--7046},
  year={2023}
}

@inproceedings{liang2025look,
  title={Look inside for more: Internal spatial modality perception for 3D anomaly detection},
  author={Liang, Hanzhe and Xie, Guoyang and Hou, Chengbin and Wang, Bingshu and Gao, Can and Wang, Jinbao},
  booktitle={Proceedings of the AAAI Conference on Artificial Intelligence},
  volume={39},
  number={5},
  pages={5146--5154},
  year={2025}
}

@inproceedings{bae2023pni,
  title={PNI: Industrial anomaly detection using position and neighborhood information},
  author={Bae, Jaehyeok and Lee, Jae-Han and Kim, Seyun},
  booktitle={Proceedings of the IEEE/CVF International Conference on Computer Vision},
  pages={6373--6383},
  year={2023}
}

@inproceedings{lei2023pyramidflow,
  title={Pyramidflow: High-resolution defect contrastive localization using pyramid normalizing flow},
  author={Lei, Jiarui and Hu, Xiaobo and Wang, Yue and Liu, Dong},
  booktitle={Proceedings of the IEEE/CVF conference on computer vision and pattern recognition},
  pages={14143--14152},
  year={2023}
}

@inproceedings{li2024towards,
  title={Towards scalable 3d anomaly detection and localization: A benchmark via 3d anomaly synthesis and a self-supervised learning network},
  author={Li, Wenqiao and Xu, Xiaohao and Gu, Yao and Zheng, Bozhong and Gao, Shenghua and Wu, Yingna},
  booktitle={Proceedings of the IEEE/CVF conference on computer vision and pattern recognition},
  pages={22207--22216},
  year={2024}
}

@inproceedings{zhou2024r3d,
  title={R3d-ad: Reconstruction via diffusion for 3d anomaly detection},
  author={Zhou, Zheyuan and Wang, Le and Fang, Naiyu and Wang, Zili and Qiu, Lemiao and Zhang, Shuyou},
  booktitle={European conference on computer vision},
  pages={91--107},
  year={2024},
  organization={Springer}
}

@inproceedings{he2024diffusion,
  title={A diffusion-based framework for multi-class anomaly detection},
  author={He, Haoyang and Zhang, Jiangning and Chen, Hongxu and Chen, Xuhai and Li, Zhishan and Chen, Xu and Wang, Yabiao and Wang, Chengjie and Xie, Lei},
  booktitle={Proceedings of the AAAI conference on artificial intelligence},
  volume={38},
  number={8},
  pages={8472--8480},
  year={2024}
}

@inproceedings{wyatt2022anoddpm,
  title={Anoddpm: Anomaly detection with denoising diffusion probabilistic models using simplex noise},
  author={Wyatt, Julian and Leach, Adam and Schmon, Sebastian M and Willcocks, Chris G},
  booktitle={Proceedings of the IEEE/CVF conference on computer vision and pattern recognition},
  pages={650--656},
  year={2022}
}

@inproceedings{deng2022anomaly,
  title={Anomaly detection via reverse distillation from one-class embedding},
  author={Deng, Hanqiu and Li, Xingyu},
  booktitle={Proceedings of the IEEE/CVF conference on computer vision and pattern recognition},
  pages={9737--9746},
  year={2022}
}

@inproceedings{rudolph2023asymmetric,
  title={Asymmetric student-teacher networks for industrial anomaly detection},
  author={Rudolph, Marco and Wehrbein, Tom and Rosenhahn, Bodo and Wandt, Bastian},
  booktitle={Proceedings of the IEEE/CVF winter conference on applications of computer vision},
  pages={2592--2602},
  year={2023}
}

@inproceedings{roth2022towards,
  title={Towards total recall in industrial anomaly detection},
  author={Roth, Karsten and Pemula, Latha and Zepeda, Joaquin and Sch{\"o}lkopf, Bernhard and Brox, Thomas and Gehler, Peter},
  booktitle={Proceedings of the IEEE/CVF conference on computer vision and pattern recognition},
  pages={14318--14328},
  year={2022}
}

@inproceedings{fuvcka2024transfusion,
  title={Transfusion--a transparency-based diffusion model for anomaly detection},
  author={Fu{\v{c}}ka, Matic and Zavrtanik, Vitjan and Sko{\v{c}}aj, Danijel},
  booktitle={European conference on computer vision},
  pages={91--108},
  year={2024},
  organization={Springer}
}

@inproceedings{yao2024glad,
  title={Glad: Towards better reconstruction with global and local adaptive diffusion models for unsupervised anomaly detection},
  author={Yao, Hang and Liu, Ming and Yin, Zhicun and Yan, Zifei and Hong, Xiaopeng and Zuo, Wangmeng},
  booktitle={European Conference on Computer Vision},
  pages={1--17},
  year={2024},
  organization={Springer}
}

@inproceedings{zavrtanik2021draem,
  title={Draem-a discriminatively trained reconstruction embedding for surface anomaly detection},
  author={Zavrtanik, Vitjan and Kristan, Matej and Sko{\v{c}}aj, Danijel},
  booktitle={Proceedings of the IEEE/CVF international conference on computer vision},
  pages={8330--8339},
  year={2021}
}

@article{yu2025registration,
  title={Registration is a Powerful Rotation-Invariance Learner for 3D Anomaly Detection},
  author={Yu, Yuyang and Chen, Zhengwei and Xu, Xuemiao and Zhang, Lei and Yang, Haoxin and Nie, Yongwei and He, Shengfeng},
  journal={arXiv preprint arXiv:2510.16865},
  year={2025}
}

@inproceedings{ye2025po3ad,
  title={Po3ad: Predicting point offsets toward better 3d point cloud anomaly detection},
  author={Ye, Jianan and Zhao, Weiguang and Yang, Xi and Cheng, Guangliang and Huang, Kaizhu},
  booktitle={Proceedings of the Computer Vision and Pattern Recognition Conference},
  pages={1353--1362},
  year={2025}
}

@article{cheng2025mc3d,
  title={Mc3d-ad: A unified geometry-aware reconstruction model for multi-category 3d anomaly detection},
  author={Cheng, Jiayi and Gao, Can and Zhou, Jie and Wen, Jiajun and Dai, Tao and Wang, Jinbao},
  journal={arXiv preprint arXiv:2505.01969},
  year={2025}
}

@inproceedings{cheng2026towards,
  title={Towards high-resolution 3d anomaly detection: A scalable dataset and real-time framework for subtle industrial defects},
  author={Cheng, Yuqi and Sun, Yihan and Zhang, Hui and Shen, Weiming and Cao, Yunkang},
  booktitle={Proceedings of the AAAI Conference on Artificial Intelligence},
  volume={40},
  number={5},
  pages={3327--3334},
  year={2026}
}

@inproceedings{li2025multi,
  title={Multi-sensor object anomaly detection: Unifying appearance, geometry, and internal properties},
  author={Li, Wenqiao and Zheng, Bozhong and Xu, Xiaohao and Gan, Jinye and Lu, Fading and Li, Xiang and Ni, Na and Tian, Zheng and Huang, Xiaonan and Gao, Shenghua and others},
  booktitle={Proceedings of the computer vision and pattern recognition conference},
  pages={9984--9993},
  year={2025}
}

@article{bergmann2018improving,
  title={Improving unsupervised defect segmentation by applying structural similarity to autoencoders},
  author={Bergmann, Paul and L{\"o}we, Sindy and Fauser, Michael and Sattlegger, David and Steger, Carsten},
  journal={arXiv preprint arXiv:1807.02011},
  year={2018}
}

@article{hong2020latent,
  title={Latent feature decentralization loss for one-class anomaly detection},
  author={Hong, Eungi and Choe, Yoonsik},
  journal={IEEE Access},
  volume={8},
  pages={165658--165669},
  year={2020},
  publisher={IEEE}
}

@inproceedings{carrara2021combining,
  title={Combining gans and autoencoders for efficient anomaly detection},
  author={Carrara, Fabio and Amato, Giuseppe and Brombin, Luca and Falchi, Fabrizio and Gennaro, Claudio},
  booktitle={2020 25th international conference on pattern recognition (ICPR)},
  pages={3939--3946},
  year={2021},
  organization={IEEE}
}

@article{cheng2025boosting,
  title={Boosting global-local feature matching via anomaly synthesis for multi-class point cloud anomaly detection},
  author={Cheng, Yuqi and Cao, Yunkang and Wang, Dongfang and Shen, Weiming and Li, Wenlong},
  journal={IEEE Transactions on Automation Science and Engineering},
  year={2025},
  publisher={IEEE}
}

@article{cheng2025toward,
  title={Toward Zero-Shot Point Cloud Anomaly Detection: A Multiview Projection Framework},
  author={Cheng, Yuqi and Cao, Yunkang and Xie, Guoyang and Lu, Zhichao and Shen, Weiming},
  journal={IEEE Transactions on Systems, Man, and Cybernetics: Systems},
  year={2025},
  publisher={IEEE}
}
\end{document}